\documentclass[10pt,twocolumn,letterpaper]{article}

\usepackage[pagenumbers]{wacv} 

\usepackage{graphicx}
\usepackage{amsmath}
\usepackage{amssymb}
\usepackage{booktabs}
\usepackage[T1]{fontenc}

\usepackage[pagebackref,breaklinks,colorlinks]{hyperref}

\usepackage[capitalize]{cleveref}
\crefname{section}{Sec.}{Secs.}
\Crefname{section}{Section}{Sections}
\Crefname{table}{Table}{Tables}
\crefname{table}{Tab.}{Tabs.}

\def\confName{WACV}

\begin{document}

\title{ARPA: A Novel Hybrid Model for Advancing
Visual Word Disambiguation Using Large Language
Models and Transformers}

\author{
    \makebox[2in][c]{Aristi Papastavrou} \quad \makebox[1.5in][c]{Maria Lymperaiou} \quad \makebox[1.5in][c]{Giorgos Stamou} \\
    \makebox[2in][c]{\tt\small{aria.papastavrou@gmail.com}} \quad \makebox[1.5in][c]{\tt\small{marialymp@islab.ntua.gr}} \quad
    \makebox[1.5in][c]{\tt\small{gstam@cs.ntua.gr}} \\
    \\
    Artificial Intelligence and Learning Systems Laboratory, \\
    School of Electrical and Computer Engineering, \\
    National Technical University of Athens
}

\maketitle

\begin{abstract}
In the rapidly evolving fields of natural language processing and computer vision, Visual Word Sense Disambiguation (VWSD) stands as a critical, yet challenging task. 
The quest for models that can seamlessly integrate and interpret multimodal data is more pressing than ever. Imagine a system that can understand language with the depth and nuance of human cognition, while simultaneously interpreting the rich visual context of the world around it.

We present ARPA, an architecture that fuses the unparalleled contextual understanding of large language models with the advanced feature extraction capabilities of transformers, which then pass through a custom Graph Neural Network (GNN) layer to learn intricate relationships and subtle nuances within the data. This innovative architecture not only sets a new benchmark in visual word disambiguation but also introduces a versatile framework poised to transform how linguistic and visual data interact by harnessing the synergistic strengths of its components, ensuring robust performance even in the most complex disambiguation scenarios. Through a series of experiments and comparative analysis, we reveal the substantial advantages of our model, underscoring its potential to redefine standards in the field. Beyond its architectural prowess, our architecture excels through experimental enrichments, including sophisticated data augmentation and multi-modal training techniques. 

ARPA’s introduction marks a significant milestone in visual word disambiguation, offering a compelling solution that bridges the gap between linguistic and visual modalities. We invite researchers and practitioners to explore the capabilities of our model, envisioning a future where such hybrid models drive unprecedented advancements in artificial intelligence.
\end{abstract}

\section{Introduction}
Visual word disambiguation (V-WSD) \cite{b6} is a crucial task in the intersection of natural language processing (NLP) and computer vision, where the goal is to correctly retrieve the appropriate image among 10 alternatives given an ambiguous word and minimal textual context. The challenging aspect of V-WSD is that the candidate images contain some confounding instances, with some of them corresponding to other meanings of the given sense, while other candidates reflect the context without any fine-grained adjustment to the ambiguous word. 
At the same time, the limited textual context in conjunction to the lack of knowledge of the ambiguous word from the side of state-of-the-art (SoTA) models for vision-language (VL) retrieval \cite{b4, b13, b14, b22, Jia2021ScalingUV, Li2022BLIPBL} frames the full range of V-WSD challenges, which is directly expressed in the limited performance of these models. 

Despite significant advancements in V-WSD performance and accompanying exploratory challenges \cite{b1, yang-etal-2023-tam, zhang-etal-2023-srcb, kritharoula-etal-2023-large, kritharoula2023languagemodelsknowledgebases}, existing approaches often struggle to fully leverage the rich information available from both linguistic and visual data, leading to suboptimal performance.

In this work, we propose a novel advanced model architecture to combat V-WSD challenges by harnessing existing lightweight components for vision and language representations. Specifically, we independently process vision and language using transformer structures tailored for each modality, ultimately projecting them onto the same embedding space. Vision-language relationship representations are boosted by passing the vision-language embeddings in a Graph Neural Network (GNN) structure which incorporates both local and global information from the data structure. Moreover, we investigate the utilization of text and image augumentation techniques to further enhance the efficiency and robustness of the proposed approach.

Overall, our contributions are:
\begin{itemize}
    \item An effective architecture that's based on simple components like LMs, Transformers and Graphs to tackle the novel problem of V-WSD.  

    \item The incorporation of a Graph Convolutional Network (GCN) into Vision-Language (VL) architectures is a novel approach that enhances contextual understanding by explicitly capturing relationships and dependencies between image and text elements. This integration allows for more nuanced relational learning, making the model more effective for tasks like Visual Word Sense Disambiguation (V-WSD), where context plays a crucial role.
    
    \item The integration of \textbf{advanced data augmentation and multi-modal training techniques}, enhancing the model's capability to generalize and perform well in real-world scenarios.

    \item Overall, we \textbf{outperform} prior V-WSD SoTa by 15\% in accuracy and 6-8\% in MRR. 
     
\end{itemize}

\section{Related work} V-WSD is a very recent task that was introduced in SemEval 2023 \cite{b6}, lying on the intersection of VL retrieval \cite{vl-survey} and word sense disambiguation \cite{wsd-survey}. Despite large-scale pre-training, VL retrieval models are of limited performance in V-WSD \cite{b4, laion}, since they lack disambiguation capabilities in their default design.
Competitive V-WSD approaches include contrastive learning mechanisms \cite{yang-etal-2023-tam}, usage of word sense inventories and lexical databases for disambiguation \cite{zhang-etal-2023-srcb, b1, grbowiec-2023-opi} or ensembling multimodal techniques \cite{patil-etal-2023-rahul}. The usage of learning-to-rank (LTR) modules combined with knowledge enhancement for disambiguation arises as a very fast and lightweight technique that also results in advanced performance \cite{b1, kritharoula-etal-2023-large}. Large Language Models (LLMs) acting as knowledge bases are able to enrich the ambiguous input phrases with appropriate context, significantly improving baselines, given proper LLM selection and suitable prompting \cite{kritharoula2023languagemodelsknowledgebases, kritharoula-etal-2023-large}.

\section{Model architecture}
\label{sec:architecture}
Our proposed model architecture is illustrated in Figure. Denoting as $T$ the linguistic modality and as $I$ the visual one, we describe the constituents of our model in the next paragraphs.

\paragraph{Language Model (LM) layer}
The initial layer of ARPA processes textual inputs $T$ by using an LM, in the default case being \textbf{RoBERTa model} \cite{b23}, a BERT based model, trained on a significantly larger dataset and with a more effective training procedure, designed to retain most of BERT's \cite{devlin-etal-2019-bert} performance while being more efficient. Via RoBERTa, the textual inputs $T$ are mapped onto a lower-dimensional space $U$ in the form of text embeddings $U_T$. We also tried to use DistilBert\cite{b10} instead of RoBERTA as it is 40\% smaller than BERT and still retains 97\% of BERT performance. However we noticed no visible difference between the two LMs, so whichever is used works the same. As an experimentation, we integrate larger models in place of RoBERTa to evaluate performance advancements, sacrificing computational efficiency as a trade-off. 

\paragraph{Vision Transformer layer}
For visual inputs $I$, we employ the \textbf{Swin Transformer}\cite{b8}. Swin is preferred over the standard ViT \cite{Dosovitskiy2020AnII} due to its advanced hierarchical architecture, designed to compute image representations through a structure that captures both local and global visual features at various scales. The Swin Transformer’s ability to dynamically adjust to different scales of visual data ensures the extraction of rich and detailed image embeddings $U_I$, which are of pivotal importance for V-WSD.

\paragraph{Linear layers} are leveraged to decrease the dimensionality of $U_T$ and $U_I$ embeddings, aligning the feature dimensions and making the representations more manageable for further processing. Separate linear layers are utilized for the embeddings of each modality. This step is crucial for effectively combining the textual and visual information.

\paragraph{Feature Combination}
The outputs from the linear layers are concatenated to form a unified representation $U_{T, I}$ that encapsulates both textual and visual information, ensuring that the model acquires a holistic view of the multimodal representation of V-WSD data.

Consider each row \( t_i \) of the textual matrix $U_T$ that corresponds to the feature vector of the \( i \)-th text instance. Similarly, given the matrix of image features $U_I$, each row \( i_j \) corresponds to the feature vector of the \( j \)-th image instance. Assuming \( t_i \) and \( i_j \) are aligned (i.e., they belong to the same data point and \( i = j \)), the concatenation of these features for each instance can be represented as:
\[ x_k = [t_k; i_k] \]
Here, \( x_k \) is the concatenated feature vector for the \( k \)-th instance belonging in $U_{T, I}$.

\paragraph{Graph Neural Network (GNN)}
The concatenated features are fed into a custom Graph Neural Network (GNN) layer, specifically a Graph Convolutional Network (GCN) \cite{b11}, which allows for accurately modelling relationships and interactions between multimodal features.

The GCN receives the concatenated text-image features $x_k \in U_{T, I}$, outputting for each node \( k \):
\[ h_k^{(l+1)} = GCN(x_k, \{h_j^{(l)} : j \in N(k)\}) \]
where \( h_k^{(l+1)} \) is the feature vector of node \( k \) at layer \( l+1 \), \( N(k) \) represents the set of neighbors of node \( k \) in the graph, and \( \{h_j^{(l)} : j \in N(k)\} \) represents the set of feature vectors of the neighbors of node \( k \) at layer \( l \). The initial feature vectors \( h_k^{(0)} \) are the concatenated vectors \( x_k \)  .

\paragraph{Output layer}
The processed and enriched feature set is passed through an output linear layer, which reduces the concatenated feature vector to a single scalar output for each instance, followed by a sigmoid activation function to produce a probability between 0 and 1. Then a  squeeze function is used to remove any singleton dimensions from the output tensor, ensuring the final output has the correct shape for binary classification. In the context of V-WSD, this layer classifies the sense of a word based on the learned multimodal embeddings.

\begin{figure}[htbp]
\centering
\includegraphics[width=0.5\textwidth]{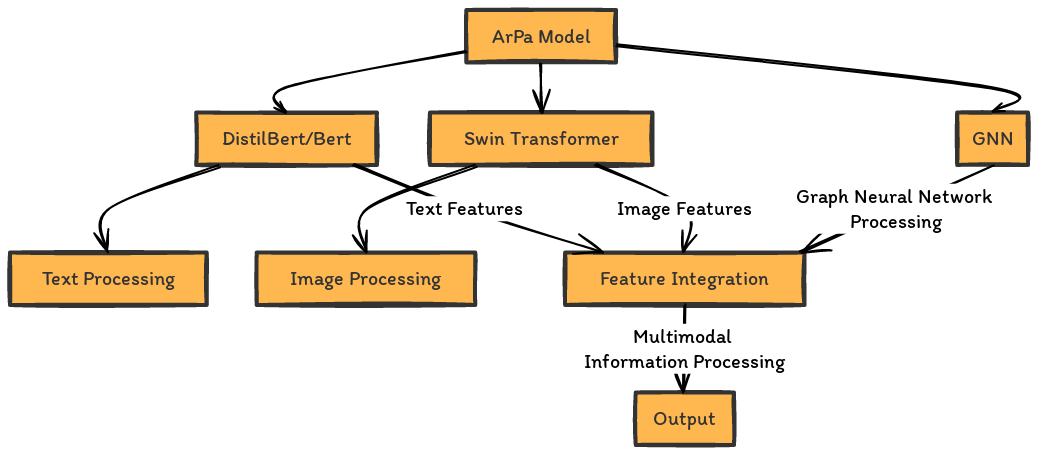}
\caption{ARPA Architecture}
\label{fig:arp_arch}
\end{figure}

\section{Enrichment Techniques}
Enhancing ARPA’s performance and robustness requires implementing and evaluating various enrichment techniques, including data augmentation and multimodal training strategies.

\paragraph{Preprocessing}is a crucial first step before any enrichment technique. Through this step, we are able to enhance our model's ability to understand and disambiguate word meanings accurately by exposing it to varied lexical contexts and aligning multimodal features effectively.
More specifically, we used WordNet, a lexical database for the English language, to retrieve possible senses (\textbf{synsets}) for each given ambiguous word, which facilitate understanding different meanings within the same context. To further increase the model's robustness, we create \textbf{lexical substitutions} by replacing target words with their synonyms based on the synsets, exposing the model to varied lexical contexts. Additionally, we utilize the \textbf{dot product similarity} between image and text embeddings $U_I, U_T$ as a crucial measure for aligning multimodal features, ensuring a cohesive integration of visual and textual information.

\subsection{Data augumentation techniques}
\label{sec:augmentation}
Data augmentation is crucial for improving the generalization ability of ARPA, targeting both linguistic and visual augumentations.

To enhance the textual data, several augmentation techniques were implemented:
\begin{itemize}
      
    \item \textbf{Random Insertion} \cite{b20}: Randomly selected words (mostly nouns, adjectives and adverbs selected from WordNet) are inserted at random and different positions in the sentences. In random insertion, we don't prioritize the exact position of the words in a sentence because the primary goal is to introduce variability and diversity into the training data. This technique helps the model handle varied sentence structures.

    \item \textbf{Random Deletion} \cite{b17}: Similarly as before, words are randomly deleted from sentences, forcing the model to rely on the remaining context to understand the sentence.

    \item \textbf{Back Translation} \cite{b19}: Sentences were translated to another language (we experimented with translating from Farsi and Italian to English) and then back to the original language. This method created paraphrased versions of the original text, enhancing the model’s ability to understand different phrasings.
\end{itemize}

On the other hand, for visual data we followed a different approach \cite{b16}, where a range of augmentation techniques were applied to increase the robustness of ARPA:

\begin{itemize}
    \item \textbf{Rotation and Flipping} \cite{b24}: Images were randomly rotated and horizontally flipped. By randomly rotating images, the model learns to recognize objects from different angles, which is crucial for tasks where the orientation of objects can vary significantly in real-world scenarios. For instance, an object might appear upright, tilted, or completely inverted. Rotation augmentation forces the model to learn the invariant features of objects, regardless of their orientation, thus enhancing its ability to generalize across varied perspectives. Horizontal flipping is another powerful augmentation technique that involves creating mirror images of the original images. This helps the model become invariant to the left-right orientation of objects. In many cases, objects can appear in mirrored forms (e.g., symmetrical objects or scenes), and horizontal flipping helps the model to handle such variations. This technique made the model invariant to the orientation of the objects in the images.

    \item \textbf{Gaussian Noise}: Random noise was added to the images to make the model robust against noisy inputs. This technique ensures that the model learns to identify and focus on the essential features of an image while ignoring irrelevant information. By introducing variability in the pixel values, the model is trained to be more resilient to inconsistencies and perturbations, improving its generalization capability and performance on unseen data. This robustness is crucial for handling real-world variations and ensuring reliable predictions across different environments and conditions.
\end{itemize}

\subsection{Multi-modal Training Strategies}

To effectively combine and leverage the strengths of both textual and visual data, several multi-modal training strategies were explored.
As described in Section \ref{sec:architecture}, image and text representations were acquired via pre-trained encoders, namely RoBERTa and Swin Transformer respectively.
A \textbf{joint embedding space} \cite{b18}, \cite{b9} was developed by projecting both text and image embeddings $U_I, U_T$ after their alignment via the linear layer into a common dimensional space $U_{T, I}$. Two versions of this strategy, that we experimented with were \textbf{Early Fusion} and \textbf{Late Fusion}.

\begin{itemize}
    \item \textbf{Early Fusion}: This approach involves combining the features from $T$ and $I$ at the initial stages of the network, (right after they are initially encoded into their respective representations). By concatenating the features early, the model can jointly learn from both modalities right from the start. This early integration allows the model to capture the interactions and dependencies between $T$ and $I$ features more effectively, potentially leading to a richer and more cohesive representation.

    \item \textbf{Late Fusion}: In this approach, features from $T$ and $I$ are processed independently through separate pathways. Each modality is allowed to learn its features without interference from the other. The learned features from the $T$ and $I$ modalities are then combined at a later stage in the network. 
    Specifically, the text and image data are processed separately through their own initial and intermediate layers. They are only merged at a much later stage, typically towards the end of the network, after each modality has been independently processed and transformed into high-level features. This method enables each modality to develop its specialized representation before merging, which can be beneficial in preserving the distinct characteristics of each modality.
\end{itemize}

Furthermore, \textbf{cross-modal attention mechanisms} \cite{b14} are attempted to enable the model to focus on relevant parts of the text and images simultaneously. This approach allow the model to dynamically adjust its attention based on the context provided by both modalities.
\section{Experiments}

\paragraph{Experimental setup}
The primary dataset used for V-WSD tasks was sourced from the \textbf{SemEval 2023 Task 1 dataset} \cite{b6}. This dataset is comprised of 12869 training samples and 463 test samples where each sample contains 10 images. 
We train the model using 10 epochs and batch size 24. Our experiments are conducted using an Nvidia GeForce RTX 4070 Super.
We evaluate our results using accuracy and Mean Reciprocal Rank (MRR) on par with prior work.

\paragraph{Preprocessing} Both text and image data underwent preprocessing to optimize them for model training. Textual data were tokenized and processed using the RoBERTa tokenizer, ensuring consistency with the RoBERTa layer. In cases we substitute RoBERTa with an alternative LM, we first apply the corresponding tokenizer. Moreover, image data were resized and normalized according to the requirements of the Swin Transformer \cite{b8}. Data augmentation techniques (Section \ref{sec:augmentation}) were applied after preprocessing.

\subsection{Results}
We conducted extensive experiments to assess the performance of ARPA in V-WSD data of mixed languages (English,Farsi and Italian), in comparison to SoTA approaches for V-WSD as well as VL retriever baselines. As a VL retriever baseline we consider CLIP \cite{b4}, while SoTA V-WSD approaches involve TAM of SCNU  \cite{TAM_SCNU}, SRC - Beijing \cite{src}, OPI \cite{zywiolak}. Related results are presented in Table \ref{table:performance_comparison}. 

\begin{table}[h]
\centering \small
\begin{tabular}{l|c|c}
\hline
\textbf{Model} & \textbf{Accuracy (\%)} & \textbf{MRR (\%)} \\ \hline
CLIP (baseline) & 37.20 & 54.39  \\ 
\hline
TAM of SCNU   & 72.56 & 82.22    \\ 
SRC - Beijing & 71.83 & 80.72    \\
zywiolak (OPI)  & 70.49 & 79.80  \\ 
\hline
\textbf{ARPA}   & \textbf{84.71}  & \textbf{88.69}    \\ \hline
\end{tabular}
\caption{Performance comparison of our model vs prior  V-WSD SoTA approaches for mixed languages dataset (English, Farsi, Italian).}
\label{table:performance_comparison}
\end{table}

The ARPA model outperforms other models in the SemEval 2023 Task 1 due to several key architectural advantages.

The ARPA model demonstrates superior performance in Visual Word Sense Disambiguation (V-WSD) across multiple datasets, significantly outperforming other state-of-the-art models in this domain. Specifically, ARPA achieves higher accuracy and mean reciprocal rank (MRR) scores compared to models such as TAM of SCNU, SRC - Beijing, and zywiolak (OPI). This enhanced performance can be attributed to several key architectural advancements within ARPA.

The TAM of SCNU model, which achieves an MRR of 82.22\%, primarily utilizes traditional contrastive learning approaches. In contrast, ARPA integrates the powerful contextual capabilities of (Large) Language Models ((L)LMs) with the advanced feature extraction capabilities of the Swin Transformer. This combination enables ARPA to capture both fine-grained textual semantics and intricate visual patterns, leading to more precise word sense disambiguation.

The SRC - Beijing model, with an MRR of 80.72\%, lacks the deep multimodal integration that ARPA offers. ARPA’s use of a Graph Convolutional Network (GCN) further enhances its ability to model relational dependencies between different elements in the data, allowing for a more comprehensive contextual understanding. This capability is crucial for accurately interpreting subtle visual and textual cues, which is why ARPA achieves superior performance compared to SRC - Beijing. Similarly the zywiolak (OPI) model, achieving an MRR of 79.80\%, is also outperformed by our model as ARPA’s architecture is designed to integrate and process multimodal data more effectively, particularly through the use of GCNs, which allows it to model complex interrelationships within the data. This results in a significantly higher MRR of 88.69\%, demonstrating ARPA’s capability in handling the complexities of V-WSD tasks, especially in multilingual and multimodal scenarios.

\begin{figure}[htbp]
\centering
\includegraphics[width=0.48\textwidth]{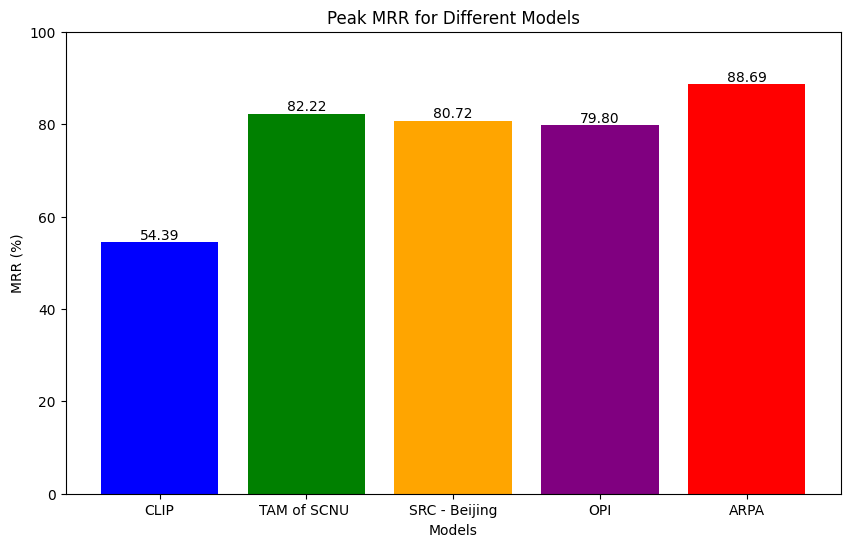}
\caption{MRR measurements for different published Models}
\label{fig:arpa_bar_mrr}
\end{figure}

\subsection{Evaluation of Enrichment Techniques}
To evaluate the impact of the enrichment techniques, the ARPA model was trained and tested with and without these techniques. The performance metrics used were accuracy and MRR. The data augmentation techniques significantly improved the model's generalization ability while the multi-modal training strategies further enhanced the model's performance. The results are summarized respectively in Tables \ref{table:augmentation_impact}, \ref{table:multimodal_impact} respectively.
\\

\begin{table}[htbp]
\centering \small
\begin{tabular}{l|c|c}
\hline
\textbf{Strategy} & \textbf{Accuracy (\%)} & \textbf{MRR (\%)} \\ \hline
No Augmentation & 59.72 & 66.66  \\ 
Text Augmentation & 72.4 & 82.13 \\ 
Image Augmentation & 61.92 & 73.68 \\ 
Both Augmentations & \textbf{82.53} & \textbf{86.24} \\ \hline
\end{tabular}
\caption{Impact of Data Augmentation on model performance.}
\label{table:augmentation_impact}
\end{table}

After applying text and image augmentation to our data, we then train the model again and applying some of the multi-modal training strategies.

\begin{table}[htbp]
\centering \small
\begin{tabular}{l|c|c}
\hline
\textbf{Strategy} & \textbf{Accuracy (\%)} & \textbf{MRR (\%)}  \\ \hline
Cross-modal Attention & 74.8 & 72 \\ 
Early Fusion & 51.58 & 50.27  \\ 
Late Fusion & \textbf{84.71}  & \textbf{88.69}  \\ \hline 
\end{tabular}
\caption{Impact of Multi-modal Training Strategies on model performance.}
\label{table:multimodal_impact}
\end{table}

\begin{figure}[htbp]
\centering
\includegraphics[width=0.5\textwidth]{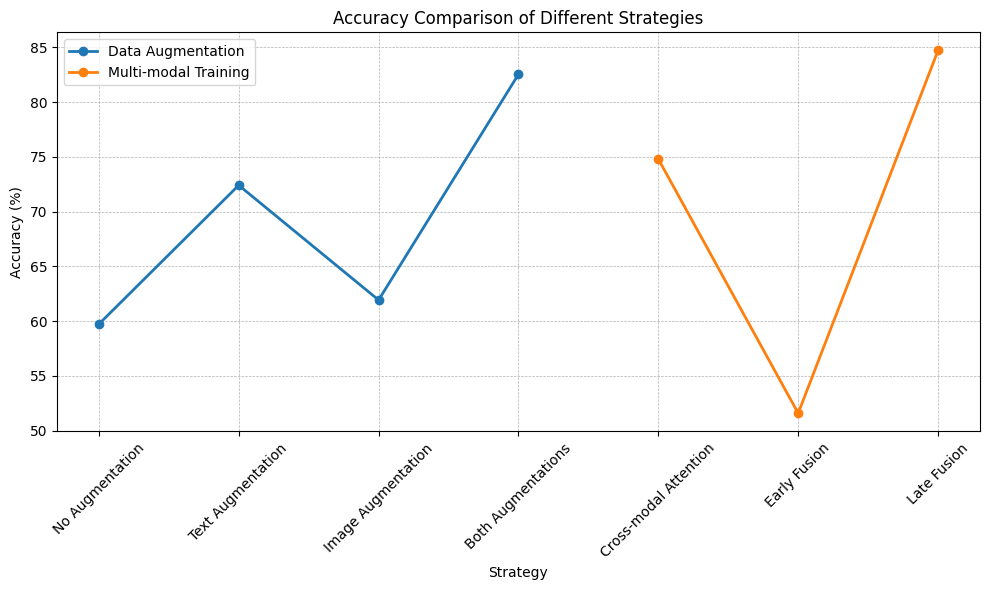}
\caption{Accuracy Comparison of Different Enrichment Techniques}
\label{fig:arpa_acc}
\end{figure}

The data augmentation techniques significantly improved the model's generalization ability when applied on both modalities. Text-only augmentation presented a non-negligible improvement on its own, while image-only augmentation only improved the overall performance by $\sim 2\%$ more than the augmentation-free baseline, suggesting that it is less effective at capturing the necessary information for robust model performance on its own.
Regarding multi-modal training strategies, results greatly vary, with a significance metric increase when the Late Fusion option is incorporated on top of the text and image augmentation.

\subsection{Architectural ablations}
The layer-wise ablation study is conducted by systematically replacing each layer of the architecture with different options to observe how these modifications affect the final results.

\paragraph{Language Model (LM) layer}
We replace the default LM (RoBERTa) with Large Language Models (LLMs), like \textbf{LLaMA} \cite{b5} (in our case we trained the LLaMA-7B: 7 billion parameters), to observe whether we would get significantly better results, since RoBERTa is a comparatively smaller model. Indeed, this alteration resulted in slightly better performance metrics, with an increase of $\sim 4\%$ in accuracy and $1-2\%$ on MRR. However, this improvement came at the cost of increased computational resources and higher model complexity compared to RoBERTa, since these models demand significantly more processing power and memory. So in order to train and evaluate our model on the dataset of the contest it would take more than a couple of hours (using the same resources as before), making the slight increase in metrics seem indifferent. However this might not be the case for more complex and bigger datasets, where the extra information and training carried by an LLM could prove more helpful than using a simple LM. 

\paragraph{Vision Transformer layer} To experimentally confirm our initial choice of Swin Transformer over the standard ViT \cite{Dosovitskiy2020AnII}, we also experiment with ViT as the encoder of the visual modality. We observe a noticeable decline in performance metrics (a decrease of $\sim$ 20-30\%), justifying our initial hypothesis that ViT lacks the necessary sophisticated mechanisms (such as the hierarchical feature representation and the local attention mechanisms) in comparison to Swin, thus struggling to achieve the same level of accuracy and robustness in visual feature extraction. 

\paragraph{Graph Neural Network (GNN)} The intuition behind employing the GCN module in order to capture more complex relations between $T$ and $I$ instead of just pipelining two models to process each stream separately is validated by removing the respective module and measure our model's performance. Indeed, we observe a noticeable decline in accuracy and MRR by $\sim 10\%$ and $\sim 17-20\%$ resepectively, illustrating the crucial contribution of the GCN towards properly comprehending the inherent semantic data interdependencies present in our multimodal representation.

\paragraph{Best overall performance combination:}Concluding the ablation study, we figured out that the best results, based purely on the accuracy and MRR, is the combination of LLaMA, Swin Transformer and Graph Convolution Network.However, considering that substituting the LM layer (RoBERTa or DistilBERT) with an LLM like LLaMA might yield a small improvement in our metrics but would significantly increase the training time and resource requirements, we believe this is not the most optimal pipeline. Since the same result can be acquired with intensive fine-tuning (using grid search or other methods), we suggest that the best performance combination is \textbf{RoBERTa as the LM layer, Swin Transformer gor the Vision Transformer layer and finally GCN}, while applying the preprocessing techniques discussed in sections \ref{table:augmentation_impact} and \ref{table:multimodal_impact}. This way we end up with a faster, equally efficient and more light-weight model.

\begin{table}[htbp]
\centering \small
\begin{tabular}{l|c|c}
\hline
\textbf{Layer} & \textbf{Accuracy (\%)} & \textbf{MRR (\%)}  \\ \hline
LM Layer (RoBERTa|DistilBert)& \textbf{88.44} & \textbf{89.31} \\ 
Vision Transformer Layer  & 62.50 & 56.94  \\ 
Graph Neural Network Layer & 75.02  & 66.66 \\ \hline 
\end{tabular}
\caption{Impact of the changes discussed in the ablation study. In each row we calculate the Accuracy and MRR of our model after keeping every layer like presented in section \ref{sec:architecture}, and changing the layer under investigation.}
\label{table:ablation_impact}
\end{table}

\section{Conclusion}
 ARPA stands as a significant advancement in the field of visual word sense disambiguation, effectively leveraging the strengths of large language models (LLMs), the Swin Transformer, and Graph Neural Networks (GNNs) to achieve superior performance. Our extensive evaluations demonstrate that our architecture not only achieves state-of-the-art results but also shows remarkable robustness and adaptability across a variety of tasks. Although the architectural pipeline resembles that of a dual-stream vision-language models, our model achieves a much more enhanced performance by employing advanced fusion techniques, including a Graph Neural Network (GCN) layer, which allows for deeper integration and context-aware processing of text and image features. This, combined with high-quality feature extraction using the Swin Transformer and simple LMs, enable ARPA to capture complex relationships and nuances within the data, leading to superior performance in VL tasks.
Additionally, the model’s ability to integrate and process multimodal inputs enables it to handle complex scenarios where text and visual data interplay, further enhancing its applicability and effectiveness. This holistic approach to visual word sense disambiguation sets a new benchmark in the field, paving the way for more sophisticated and versatile applications in natural language processing and computer vision. 

There is of course space for improvement. The training and inference processes of our model require substantial computational resources, including high-performance GPUs for handling massive loads of data. This requirement can potentially be a barrier for users with limited access to such
resources.

As we move forward, we aim to enhance the efficiency of our model, ensuring it maintains its state-of-the-art performance while becoming less time-consuming. Our goal is to optimize the model’s computational requirements, streamlining both training and inference processes. By refining our architecture and leveraging cutting-edge techniques, we plan to optimise the use of resources without compromising its exceptional accuracy and robustness. This will make ARPA more accessible and practical for a wider range of applications, further solidifying its position as a leading model in visual word sense disambiguation tasks.

{\small
\bibliographystyle{ieee_fullname}
\bibliography{ARPA_main}
}

\end{document}